\documentclass[conference]{IEEEtran}

\usepackage[pdftex]{graphicx}
\graphicspath{{figures}}
\usepackage{tikz,graphics,color,fullpage,float,epsf,caption,subcaption}
\usepackage{hyperref}
\usepackage{glossaries}
\usepackage{glossaries-prefix}
\usepackage{siunitx}
\usepackage{balance}

\usepackage{url}
\usepackage{blindtext}
\usepackage{xcolor}

\newacronym{AI}{AI}{Artificial Intelligence}
\newacronym{AEB}{AEB}{Autonomous Emergency Braking}
\newacronym[
    longplural={Automated Driving Systems},
    shortplural={ADS}
    ]
    {ADS}{ADS}{Automated Driving System}
\newacronym[
    longplural={Advanced Driver Assistance Systems},
    shortplural={ADAS}
    ]
    {ADAS}{ADAS}{Advanced Driver Assistance System}
\newacronym[
    longplural={Advanced Driver Assistance Systems / Automated Driving},
    shortplural={ADASs/AD}
    ]
    {ADAS/AD}{ADAS/AD}{Advanced Driver Assistance System / Automated Driving}
\newacronym{CSP}{CSP}{Characteristic Stripe Pattern}
\newacronym{GDPR}{GDPR}{General Data Protection Regulation}
\newacronym{ML}{ML}{Machine Learning}
\newacronym{ODD}{ODD}{Operational Design Domain}
\newacronym[
    longplural={Original Equipment Manufacturers},
    shortplural={OEMs}
    ]
    {OEM}{OEM}{Original Equipment Manufacturer}
\newacronym[
    prefixfirst={a\ },
    prefix={an\ }
    ]
    {SFC}{SFC}{Space-Filling Curve}
\newacronym{VV}{V\&V}{validation \& verification}
\newacronym{ZEBRA}{ZEBRA}{Z-order Curve-based Event Retrieval Approach}

\begin{document}

\title{Systematic Evaluation of Applying Space-Filling Curves to Automotive Maneuver Detection}

\author{\IEEEauthorblockN{Christian Berger$^1$, Beatriz Cabrero-Daniel$^1$, M.~Cagri Kaya$^2$, Maryam Esmaeili Darestani$^1$, Hannah Shiels$^1$}\\
\IEEEauthorblockA{
$^1$\textit{University of Gothenburg} and $^2$\textit{Chalmers University of Technology}\\
Gothenburg, Sweden \\
\{christian.berger,beatriz.cabrero-daniel\}@gu.se, cagri.kaya@chalmers.se}
}

\markboth{Journal of \LaTeX\ Class Files,~Vol.~14, No.~8, August~2015}%
{Shell \MakeLowercase{\textit{et al.}}: Bare Demo of IEEEtran.cls for IEEE Journals}

\maketitle

\begin{abstract}
Identifying driving maneuvers plays an essential role on-board vehicles to monitor driving and driver states, as well as off-board to train and evaluate machine learning algorithms for automated driving for example. Maneuvers can be characterized by vehicle kinematics or data from its surroundings including other traffic participants. Extracting relevant maneuvers therefore requires analyzing time-series of (i) structured, multi-dimensional kinematic data, and (ii) unstructured, large data samples for video, radar, or LiDAR sensors. However, such data analysis requires scalable and computationally efficient approaches, especially for non-annotated data. In this paper, we are presenting a maneuver detection approach based on two variants of space-filling curves (Z-order and Hilbert) to detect maneuvers when passing roundabouts that do not use GPS data. We systematically evaluate their respective performance by including permutations of selections of kinematic signals at varying frequencies and compare them with two alternative baselines: All manually identified roundabouts, and roundabouts that are marked by geofences. We find that encoding just longitudinal and lateral accelerations sampled at 10\,Hz using a Hilbert space-filling curve is already successfully identifying roundabout maneuvers, which allows to avoid the use of potentially sensitive signals such as GPS locations to comply with data protection and privacy regulations like GDPR.
\end{abstract}


\IEEEpeerreviewmaketitle

\section{Introduction} \label{sec:introduction}

The development and evaluation of \glspl{ADAS} and \glspl{ADS} is a data-driven process that covers systematic and reproducible \gls{VV} using closed-loop simulations, controlled experimentation at confined test sites using prototypical vehicles, and field data collections with larger vehicle fleets for open-loop replay or scenario reconstruction. While virtual environments provide systematic and cost-effective annotations by design, the degree of fidelity in tests using real sensors in real scenarios is still higher, as subtle effects that may not be part of the models used for the virtual sensors or the environment model can be captured. In addition, a much broader variety of situations can be potentially recorded that would require too much effort to be modeled manually in virtual test environments.


Large-scale field data collections apparently have a greater value during the system development and for \gls{VV} because of the representativeness of the captured scenarios and because of the data originates from the real sensor configuration. Furthermore, such field data can grow quickly by equipping hundreds of vehicles with data loggers that stream their data to a centralized storage, where it can be analyzed further. While the data collection itself can be conducted rather cost-effectively, the actual data processing afterwards to make use of the value of such data is becoming a growing challenge: Firstly, field data is usually non-annotated right after collection and relevant labels need to be added and verified (semi-)automatically before queries for maneuvers are possible; secondly, data privacy and protection regulations such as \gls{GDPR} may not only require the anonymization of video data (for instance, blurring faces or license plates) but also potentially the removal of signal types such as GPS locations of the vehicle that collected the data to prevent the identification of individuals later.

The removal of certain signal types, though, may render the implementation and use of some maneuver detection and querying approaches infeasible. For instance, finding maneuvers when a vehicle is passing through a roundabout by defining geofences and checking when a vehicle is entering such geofences of interest would require potentially \gls{GDPR}-sensitive information. Alternatively, adding labels for entering/leaving roundabouts at the time of collecting data would require the existence of a taxonomy of scenarios to label upfront, and added elements in such a taxonomy would then only be applicable to newly collected data but not to existing data from the past.

\vspace{0.15cm}
\noindent\fbox{%
\begin{minipage}{.96\columnwidth}
\textbf{Motivation:} There is a need for an approach to detect maneuvers and events that (a) does not require the definition and maintenance of an annotation taxonomy before data is collected, (b) does not depend on \\(semi-)automatically added labels, and (c) is scalable to cope with growing amounts of data. 
\end{minipage}
}\\

Berger and Birkemeyer (cf.~\cite{ZEBRA}) introduced using a Z-order \gls{SFC} as a novel approach to explore automotive data in a computationally efficient way for event detection. In this paper, we are conducting a qualitative experiment with two different types of \glspl{SFC}, namely Z-order and Hilbert curves, to answer the following research questions:



\vspace{0.15cm}
\noindent\fbox{%
\begin{minipage}{.96\columnwidth}
\begin{description}
    \item[RQ-1:] To what extent can Z-order and Hilbert \gls{SFC} be exploited to identify automotive maneuvers in an efficient way without relying on \gls{GDPR}-sensitive data?
    \item[RQ-2:] How do different experiment configurations (e.g., used signals, sampling frequency, etc.) compare to manual annotations of automotive maneuvers?
\end{description}
\end{minipage}
}\\


We limit our systematic experiment on the detection of driving through roundabouts with a passenger vehicle (Volvo XC90). While the parametrization is adjusted for the experimental vehicle, the fundamental concept is transferable to other vehicles and scenarios. However, the best-performing combination of kinematic signals, though, may be affected by the chosen automotive maneuver in our experiments and may differ for other maneuvers and/or geographic regions with different lane layouts. 

The remainder of the paper is structured as follows: In Sec.~\ref{sec:relatedwork}, we introduce the ideas behind \glspl{SFC} and describe related works. We describe the methodology for our study in Sec.~\ref{sec:methodology} and present the results in Sec.~\ref{sec:results}. We discuss and analyze our results in Sec.~\ref{sec:analysis-discussion} and conclude our work in Sec.~\ref{sec:conclusion-futurework}.


\section{Related Work}
\label{sec:relatedwork}

One fundamental issue with event detection based on kinematic signals originates from their multi-dimensionality, which is posing a computational challenge at scale. Brute-force-based approaches like the one suggested by Perez et al.~\cite{PEREZ201710} are on the one hand traceable and allow to understand contributing factors for a positive hit, but they scale proportionally with a data set's size on the other hand and hence, are not practical for large data sets.

Alternative approaches that explicitly address multi-dimensionality are based on \glspl{SFC}. An \gls{SFC} enables a single-dimensional representation of multi-dimensional data (cf.~Bader, \cite{Bad13}) and hence, effectively allows for dimensionality reduction. Intuitively, an \gls{SFC} passes through each data point from a multi-dimensional space in a recursive or repetitive manner. Certain types of \glspl{SFC} even preserve locality properties, ie., data points that are close to each other in the multi-dimensional data space will be located near to each after being mapped onto an \gls{SFC}. This property in turn can be exploited for efficient data retrieval as operations on an \gls{SFC} are usually faster and computationally cheaper on single dimensions than on multiple dimensions simultaneously. \glspl{SFC} have been explored by Hulbert et al.~\cite{hulbert2016} for example who evaluated the performance of \glspl{SFC} on large data. Similarly, Dai et al.~\cite{dai2022clustering} compared both, Z-order and Hilbert curves applied to two-dimensional data, and Moon et al.~\cite{moon2001analysis} investigated the Hilbert \gls{SFC} for indexing multi-dimensional data that can be used for data queries.

There are also complementary studies that employ \glspl{SFC} for event detection across diverse application fields. For instance, Safia et al.~\cite{distributed-2019} propose an algorithm to detect environmental events like air pollution that can be used for monitoring by sensor networks. Liu et al.~\cite{hgst-2023} suggest a method called HGST that uses Hilbert curves to improve the efficiency of spatiotemporal range queries. The method was tested on a real taxi trajectory data set. Nair et al.~\cite{nair-2017} present their study on dynamic planning of evasive maneuvers in an environment, which represents obstacles and holes by using Hilbert curves.

\begin{figure}
    \centering
    \begin{subfigure}{0.5\textwidth}
        \centering
        \includegraphics[scale=0.3]{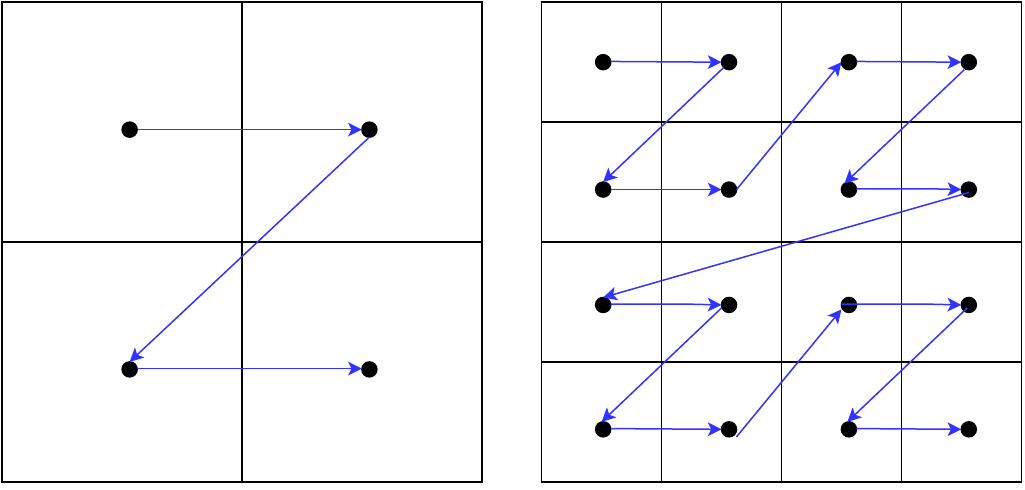}
        \caption{Traversing a two dimensional space using a Z-order curve; for the technical implementation, multi-dimensional values are encoded into so called Morton codes (cf.~Morton, \cite{Mor66}).}
    \label{fig:z-curve}
    \end{subfigure}

    \begin{subfigure}{0.5\textwidth}
        \centering
        \includegraphics[scale=0.3]{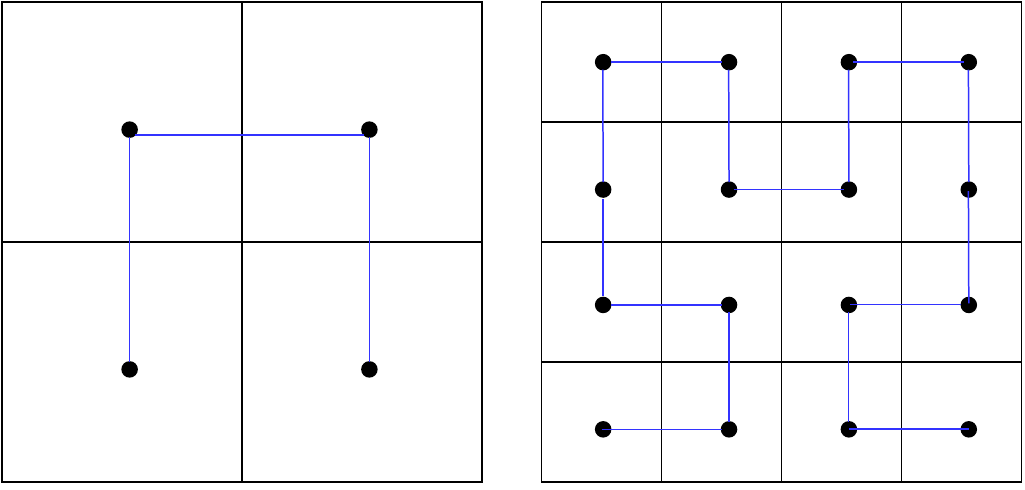}
        \caption{Traversing a two dimensional space using a Hilbert curve.}
        \label{fig:hilbert-curve}
    \end{subfigure}
    \caption{Illustrations for recursive construction of Z-order (a) and Hilbert (b) curves to traverse a two dimensional space}
    \label{fig:my_label}
\end{figure}

\begin{figure*}
     \centering
     \begin{subfigure}[b]{0.4\textwidth}
         \centering
         \includegraphics[scale=0.25]{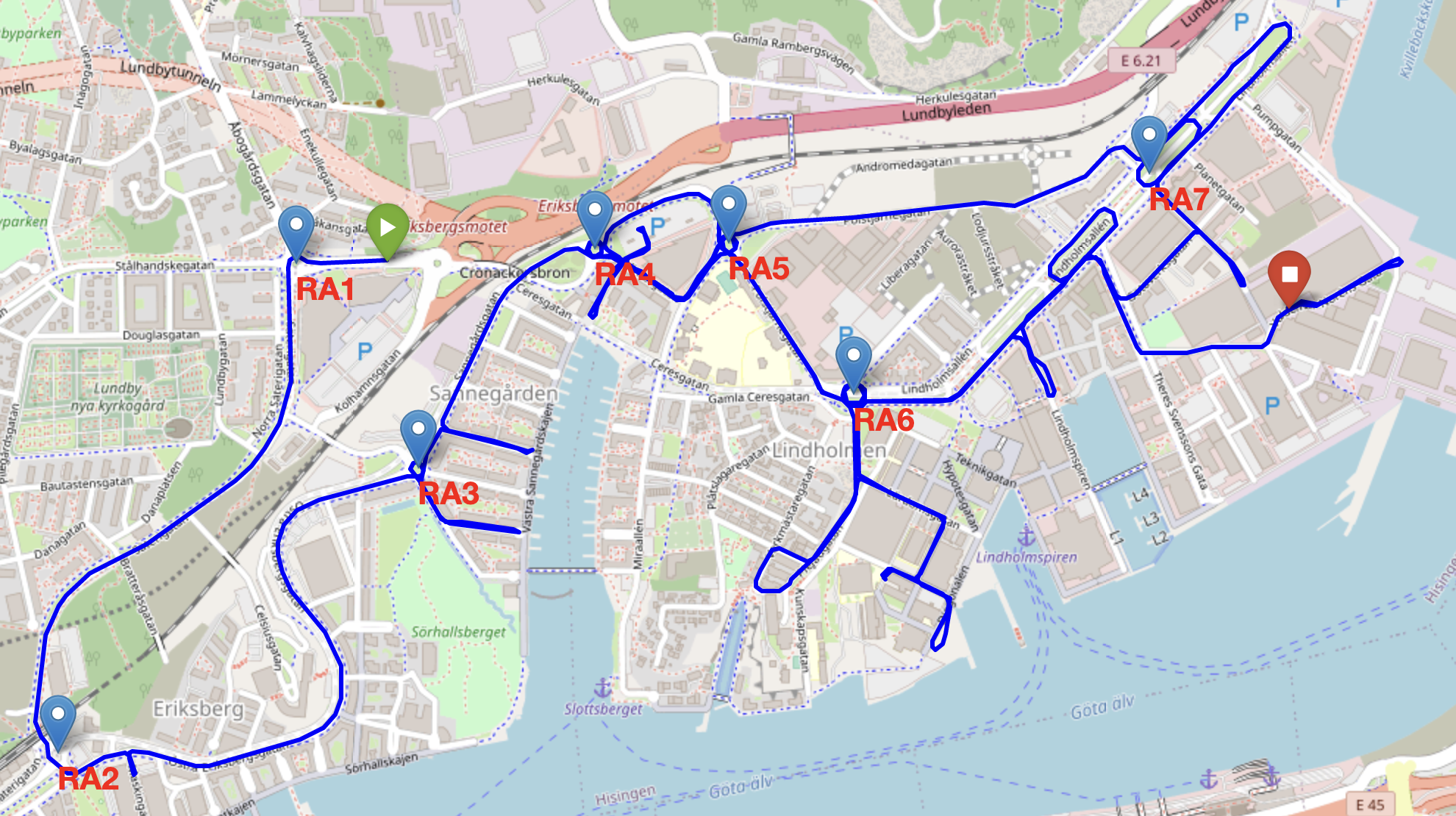}
         \caption{OpenStreetMap overlay of the trip used for experiments.}
         \label{fig:y equals x}
     \end{subfigure}
     \hfill
     \begin{subfigure}[b]{0.45\textwidth}
         \centering
            \begin{minipage}{0.5\textwidth}
            Locations of roundabouts: \\
            \vspace{0.01cm}
            \begin{tabular}{l|c|c}
            & latitude & longitude \\
            \hline
            $\text{RA1}$ & $57.70974$ & $11.91848$ \\
            \hline
            $\text{RA2}$ & $57.70207$ & $11.91152$ \\
            \hline
            $\text{RA3}$ & $57.70655$ & $11.92202$ \\
            \hline
            $\text{RA4}$ & $57.70998$ & $11.92721$ \\
            \hline
            $\text{RA5}$ & $57.71007$ & $11.93112$ \\
            \hline
            $\text{RA6}$ & $57.70771$ & $11.93477$ \\
            \hline
            $\text{RA7}$ & $57.71113$ & $11.94337$
            \end{tabular}
            \end{minipage} 
            \caption{Locations of the roundabouts.}
         \label{fig:three sin x}
     \end{subfigure}
        \caption{Overview of the trip in an inner-city area in Gothenburg, Sweden. The length of the trip is about \SI{14}{km} and took ca.~\SI{56}{min}. During the trip, 7 different roundabouts (denoted as $\text{RA} x$) were visited and some thereof even multiple times.}
        \label{fig:trip-overview}
\end{figure*}

We introduced the novel pattern detection approach named \gls{ZEBRA} based on Z-order curves that we exploited for computationally efficient event detection in the automotive context (cf.~Berger and Birkemeyer, \cite{ZEBRA}). We leveraged a Z-order \gls{SFC} to reduce multi-dimensional data into their corresponding single-dimensional representations (vertical stripes) resulting in \glspl{CSP} as shown in Fig.~\ref{fig:accelXY_CSP-Morton}. We showed that the temporal occurrence, distribution, and spread of such stripes correlate with certain maneuvers in the multi-dimensional space. The locality preservation property of \glspl{SFC} in combination with operating just on a single dimension allowed us to design computationally efficient operations to detect maneuvers. We showed in \cite{ZEBRA} that maneuver detection can be conducted orders of magnitude faster in comparison to directly processing time-series data in the original data space. The key idea behind this performance gain is motivated by the fact that the application context restricts the value range of the underlying \gls{SFC}, on which operations like inserting or looking up entries can be realized efficiently as a consequence.

In this study, we are using two types of \glspl{SFC}, namely Z-order and Hilbert curves: A Z-order curve is exhibiting a Z-like pattern when repetitively traversing a multi-dimensional space as shown in Fig.~\ref{fig:z-curve}; we chose the Z-order curve as the mapping from the multi-dimensional space to the single-dimensional representation can be computed efficiently. A Hilbert curve is a recursive traversal through a multi-dimensional data space and its pattern is depicted in Fig.~\ref{fig:hilbert-curve}; we selected the Hilbert curve as it exhibits even better locality preservation (cf.~Bader, \cite{Bad13}) for data points that are in close proximity to each other.


\section{Methodology}
\label{sec:methodology}

\begin{table}[b!]
\centering
    \begin{tabular}{|p{2.6cm}|c|c|c|c|c|} 
    \hline
        \textbf{signal} & \SI{5}{\hertz} & \SI{10}{\hertz} & \SI{20}{\hertz} & \SI{50}{\hertz} & \SI{100}{\hertz}\\ 
        \hline
			GPS position & - & - & $\bullet$ & - & -\\
			heading, $\sigma, [\text{rad}]$ & $\bullet$ & $\bullet$ & $\bullet$ & - & -\\
        \hline
			$v, [m/s]$ & $\bullet$ & $\bullet$ & $\bullet$ & $\bullet$ & $\bullet$ \\
        \hline
			steering wheel angle, $\delta_S, [\text{rad}]$ & $\bullet$ & $\bullet$ & - & $\bullet$ & -  \\
        \hline
			lateral acceleration, $a_x, [m/s^2]$ & $\bullet$ & $\bullet$ & $\bullet$ & $\bullet$ & $\bullet$ \\
			longitudinal acceleration, $a_y, [m/s^2]$ & $\bullet$ & $\bullet$ & $\bullet$ & $\bullet$ & $\bullet$ \\
			vertical acceleration, $a_z, [m/s^2]$ & $\bullet$ & $\bullet$ & $\bullet$ & $\bullet$ & $\bullet$ \\
        \hline
			yaw rate, $\Dot{\Psi}, [\text{rad}/s]$ & $\bullet$ & $\bullet$ & $\bullet$ & $\bullet$ & $\bullet$ \\
			pitch rate, $\Dot{\Theta}, [\text{rad}/s]$ & $\bullet$ & $\bullet$ & $\bullet$ & $\bullet$ & $\bullet$ \\
			roll rate, $\Dot{\Phi}, [\text{rad}/s]$ & $\bullet$ & $\bullet$ & $\bullet$ & $\bullet$ & $\bullet$ \\
        \hline
    \end{tabular}
\caption{Signals and their respective frequencies, captured at their highest available frequency on the Volvo XC90 experimental platform and down-sampled for the experiments.\label{tab:data-snowfox}}
\end{table}

The goal of the study is to compare the performance (in terms of precision, accuracy, and recall) of maneuver detectors to find potential passing maneuvers when driving through a roundabout by exploiting multi-dimensional kinematic signals as input for the \glspl{SFC}. We use a Volvo XC90 from the vehicle laboratory Chalmers Revere that has been instrumented with an Applanix GPS/IMU system to obtain a baseline reference. In addition, on-board vehicle data as listed in Tab.~\ref{tab:data-snowfox} is accessible via CAN for data logging. The vehicle was used for a manual data collection of about \SI{56}{min} for a trip of \SI{14}{km} in an urban area within Gothenburg, Sweden as shown in Fig.~\ref{fig:trip-overview}. The trip covered in total 17 roundabout scenarios by visiting 7 different roundabouts on the route. The baseline for the experiments as described in the following is defined two-fold: (a) manual annotations were added to extract the time-points when entering a roundabout, information about roundabout layout (ie., number of exits and chosen exit), and the time-points when leaving the roundabout, and (b) by defining geofences around all possible roundabouts along the trip as marked in Fig.~\ref{fig:trip-overview}.

\begin{figure*}[h!]
    \centering
    \begin{subfigure}[b]{0.49\textwidth}
         \centering
         \includegraphics[width=1.0\linewidth]{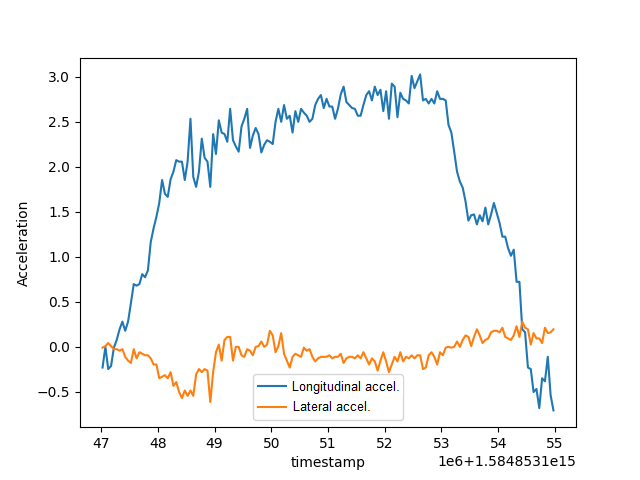}
         \caption{Lateral and longitudinal\\ accelerations over time.}
         \label{fig:accelXY_CSP-onlyaccelerations}
     \end{subfigure}
     \begin{subfigure}[b]{0.49\textwidth}
         \centering
         \includegraphics[width=1.0\linewidth]{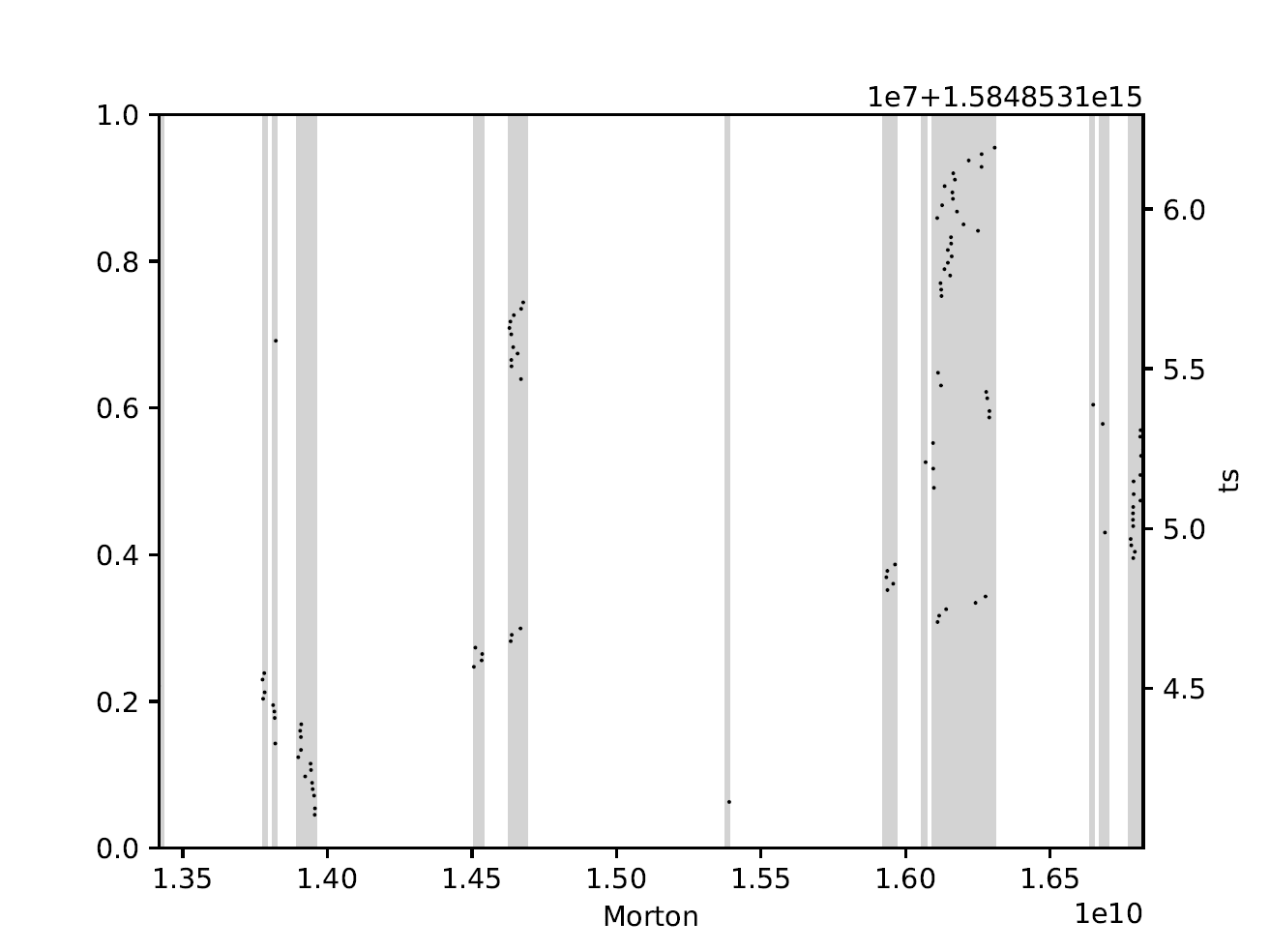}
         \caption{Corresponding Z-order curve stripes (lightgray), broken down to their corresponding occurrence over time (black dots).}
         \label{fig:accelXY_CSP-Morton}
     \end{subfigure}
    \caption{Lateral and longitudinal accelerations for passing through reference roundabout $RA_1$, which is used to design the \gls{SFC}-curves based maneuver detector (left) and the corresponding stripes after mapping to a Z-order curve based on Morton codes (right). The light-gray stripes denote the presence of a \gls{SFC}-value mapped from a multi-dimensional data tuple ($a_x$, $a_y$) with $1$, and $0$ otherwise (left Y-axis); the black dots on top of the light-gray stripes denote their respective occurrence over time (right Y-axis).
    \label{fig:accelXY_CSP}}
\end{figure*}

The geofences serve as input for an automatic GPS position-based maneuver detector that encodes the \gls{GDPR}-sensitive vehicle location onto an \gls{SFC} curve as follows: The two dimensions (latitude, longitude) were converted to single-dimensional Morton codes (cf.~Morton, \cite{Mor66} and Bader, \cite{Bad13}) resulting in a positive integer number. This number was then used as sorted index to the corresponding time $t$ of a GPS location resulting in a tuple $(\text{\emph{Morton}}(\text{latitude}_{\text{pos}},\text{longitude}_{\text{pos}}),t)$. 

The maneuver look-up is then a range query where the bottom/left and top/right GPS location of a geofence are the lower and upper limits ($\text{\emph{Morton}}(\text{latitude}_\text{BL},\text{longitude}_\text{BL})$, $\text{\emph{Morton}}(\text{latitude}_\text{TR},\text{longitude}_\text{TR})$) on the single-dimensional space. All Morton indices between such lower and upper limits correspond to potential candidate locations where the vehicle potentially was passing through a roundabout. Finally, all such potential pairs $(x,t_x)$, with $x$ between the lower/upper limits as returned from the range query, are converted back to the original tuple (latitude, longitude) and filtered subsequently using a point-in-polygon test for the corresponding geofence to compensate for potential false-positive Morton candidates resulting from sudden ``jumps'' in the traversal of the Z-order curve through the multi-dimensional space (see Fig.~\ref{fig:z-curve}, right).

In contrast to the aforementioned maneuver detector, which is using a signal that can be considered potentially being \gls{GDPR}-sensitive, we setup a systematic experiment for different maneuver detectors based on the two variants (a) Z-order curve-based \gls{SFC} using Morton codes (abbreviated as \emph{M} in Fig.~\ref{fig:results}), and (b) Hilbert curve (abbreviated as \emph{H} in Fig.~\ref{fig:results}) to answer both research questions. Both variants of \glspl{SFC} convert multi-dimensional data samples into their respective single-dimensional representations similar to the aforementioned description for the geofence-based detector. Specific \gls{CSP} as shown in Fig.~\ref{fig:accelXY_CSP-Morton}, in which the stripes occur in dedicated ranges over time, correlate with specific automotive maneuvers in the original multi-dimensional space. We use a \gls{CSP} from a known reference maneuver when passing through a roundabout ($\text{RA}_1$) to obtain the distribution of \gls{SFC} values as shown in Fig.~\ref{fig:accelXY_CSP}. To detect subsequent passing maneuvers through roundabouts, a dataset's \gls{SFC} values are traversed to check whether respective \gls{CSP} ranges and value distributions in the stripes are within the thresholds extracted from the reference passing maneuver that we used for calibrating the maneuver detector.

We systematically created permutations as listed in Fig.~\ref{fig:results} for our experimentation using the following configuration parameters to evaluate the respective performance in finding drive-through maneuvers at roundabouts:
\begin{enumerate}
    \item Select Morton or Hilbert for \gls{SFC} conversion.
    \item Select sampling frequency between 5 and \SI{100}{\hertz}.
    \item Select combinations of minimum two signals from the set: $\{a_x, a_y, \delta_S, \Dot{\Psi}\}$.
\end{enumerate}
The motivation for the first option is to systematically evaluate what \gls{SFC} variant is showing a superior performance for detecting a passing maneuver. The second option is to evaluate how much data is essential for a successful maneuver detection; the motivation here is that data minimization is not only reducing processing time and storage usage, but also obeying the data minimization principle according to \gls{GDPR}. Finally, systematic combinations of various signals is evaluating to what degree certain kinematic parameters are influencing a successful maneuver detection.

\section{Results}
\label{sec:results}

\begin{figure*}
    \centering
    \begin{tikzpicture}
    \node at (0,0) {\includegraphics[width=.95\linewidth]{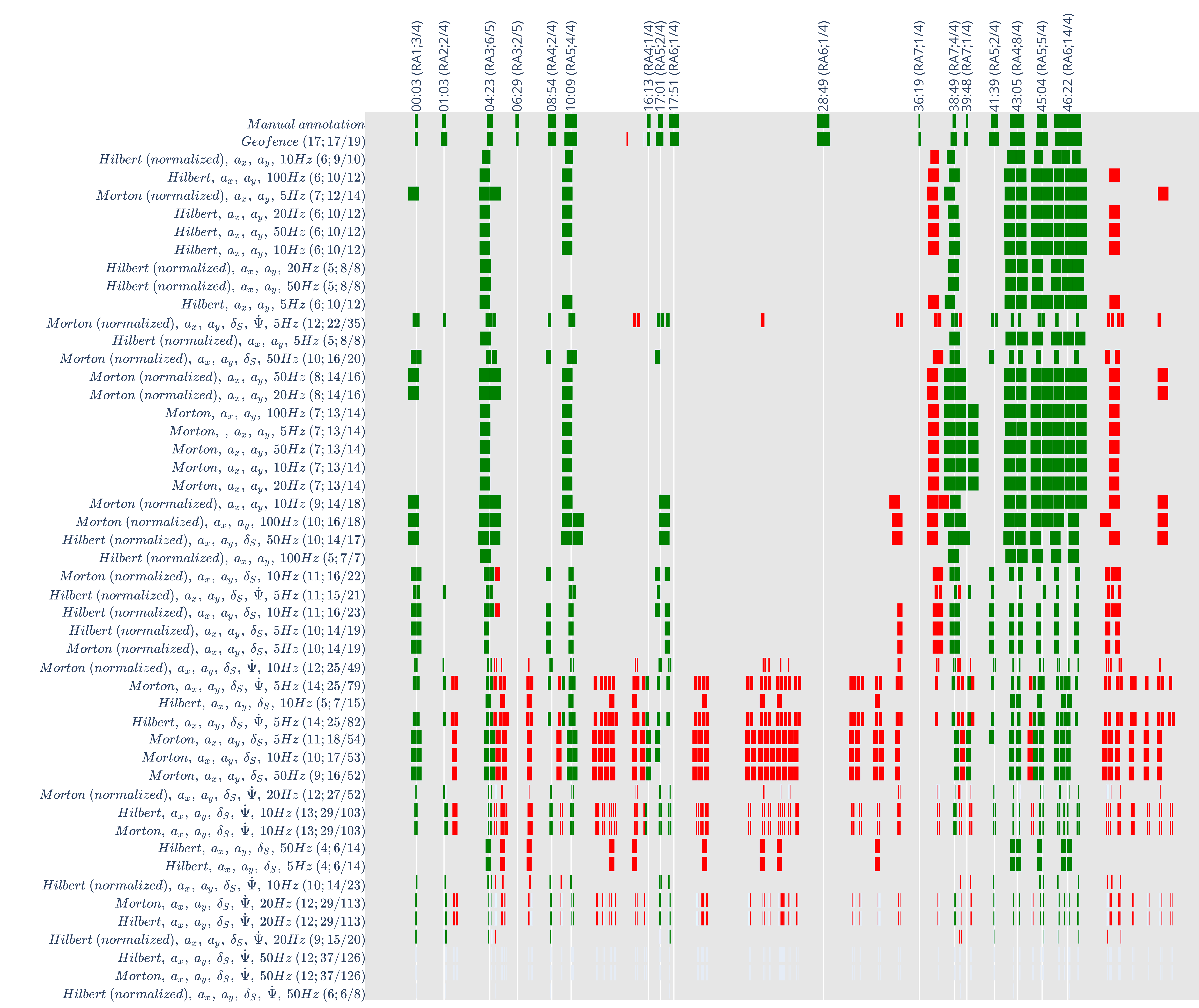}}; 
    \node[fill=cyan] at (-7.65,4.9) {$\text{Ex}_1$:};
    \node[fill=cyan] at (-8.4,2.55) {$\text{Ex}_2$:};
    \node[fill=cyan] at (-7.3,-2.7) {$\text{Ex}_3$:};
    \end{tikzpicture}
    \caption{Results from the conducted experiments ordered by F1-score that is computed based on overlapping time windows between the manually annotated ground truth and the results obtained from the particular experiments as listed in the first column; the values in the parenthesis denote the number of correctly identified, distinct passes through roundabouts (TPs) followed by the number of correctly identified passes (including duplicates) over total guesses (including FPs). The width of the red and green bars show the returned durations, respectively. The results for the last three experiments contain very short durations that cannot be displayed clearly. As column headers, we denote the time when a passing maneuver at a certain roundabout happened including the roundabout's configuration and what exit was chosen; for example, 3/4 means that the roundabout had 4 exits and the third exit was taken.
    \label{fig:results}}
\end{figure*}
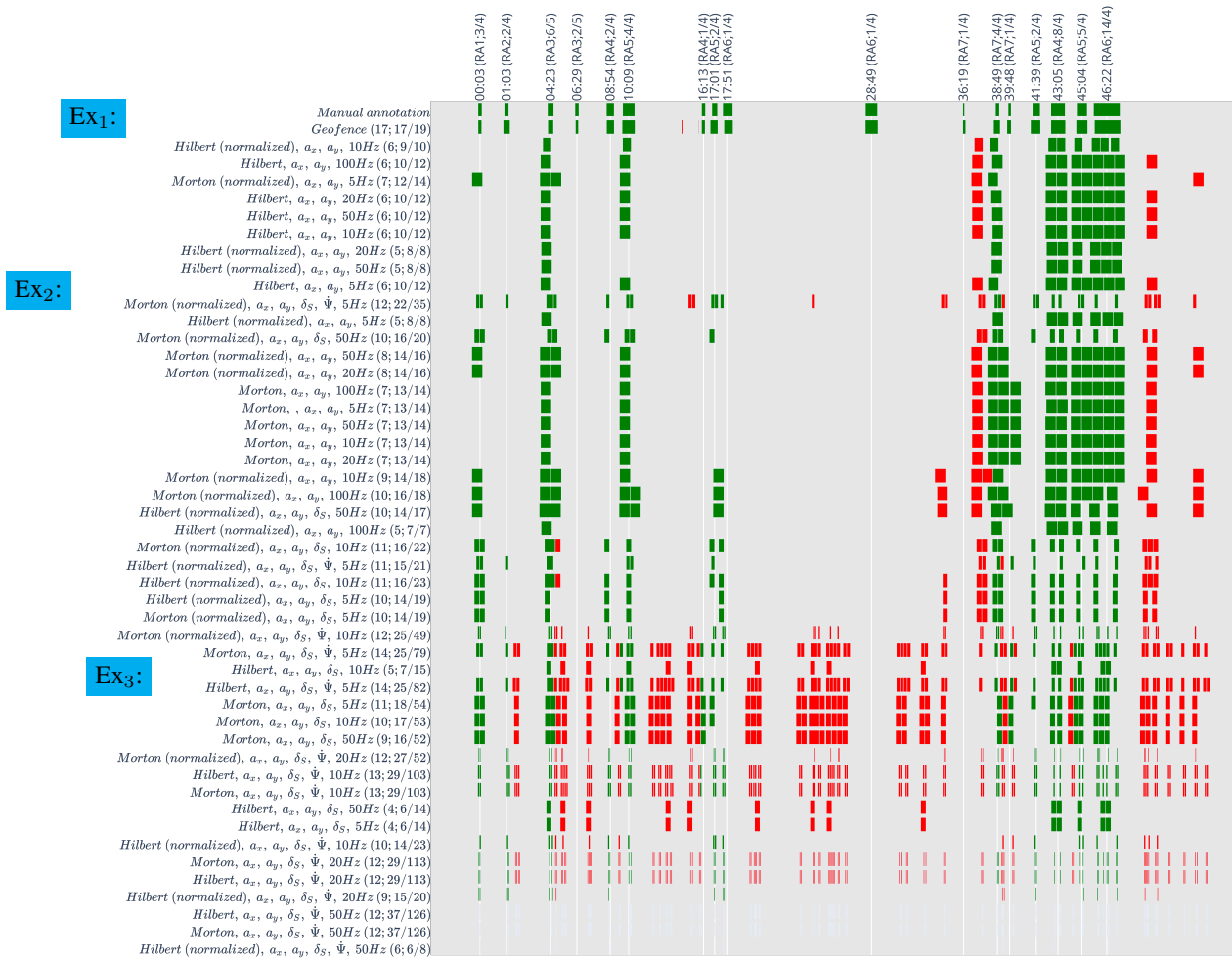

The results from our systematic experiments are aggregated and presented in Fig.~\ref{fig:results}. The first column lists the specific configuration for a single experiment, characterized by the chosen parameters: An \gls{SFC} method for encoding the multi-dimensional values, a combination of kinematic signals from the multi-dimensional space, and a data sampling frequency. Other columns depict each roundabout pass-through in the order of occurrence. Therefore, the x-axis shows the temporal dimension: the length of each box refers to the event's duration.  
The top two rows show both baselines: The manually annotated roundabout events and the results obtained from the geofence-based maneuver detector. All subsequent rows show the results from a specific configuration of an \gls{SFC}-based maneuver detector. 

The green boxes represent correctly identified passes through a roundabout (true-positive, TP) and the red boxes indicate when a maneuver detector returned time ranges from a maneuver that it believed was a passing maneuver through a roundabout but it was not (false-positive, FP). The color was assigned to each box by conducting a binary test whether the time range returned by a detector is partially overlapping with the manual annotations. Thus, green indicates an overlap, and red no overlap. In our experiments, we consider a minimal overlap being sufficient for a TP hit. We determine such overlapping time ranges by calculating the \emph{intersection-over-union} ($\text{IoU}$) of the manually annotated ground truth with the returned time ranges of an experiment. Each finding is set to green (a TP) if it has an $\text{IoU}$ value greater than $0$. In this setting, a 1-second-long overlap with a passing maneuver, a 100\% overlap with a passing maneuver, or even ten different short detections that overlap with a passing maneuver are all considered TPs, which we refer to as \textit{Boolean-identified passes}.

Alternatively, successful passes can also be defined as the amount of intersecting seconds, FPs are then defined as the detected seconds without the intersecting seconds, false-negative (FN) is defined as the ground truth time windows without the intersections, and true-negative (TN) is the remaining time (neither passing in the ground truth nor according to a detector). Fig.~\ref{fig:results} is ordered using these time-based definitions by the F1-score of the respective experiment because we prefer configurations for maneuver detectors that find as many TPs as possible without being penalized by too many FPs; intuitively speaking, we want as many green ranges for the costs of the least red ranges as possible. This is also reflected by having the manual annotations at the top (only TPs with no FPs), followed by the geofence-based detector that is using \gls{GDPR}-sensitive signals. It can be noted, though, that while this detector correctly identifies all passing maneuvers as expected, the dimensioning of the geofences may unintentionally reach into nearby streets resulting in FPs (two in our experiment).   

Based on the analysis of the results, we choose three exemplary experiment configurations with different success rates and discuss their strengths: (i) Hilbert with $\{a_x,a_y,\SI{10}{\hertz}\}$, to which we will refer as experiment $\text{Ex}_1$ from now on; (ii) Morton with $\{a_x,a_y,\delta_S,\dot{\Psi},\SI{5}{\hertz}\}$, here forth $\text{Ex}_2$; and (iii) Morton with $\{a_x,a_y,\dot{\Psi},\SI{5}{\hertz}\}$, here forth $\text{Ex}_3$. The first two configurations use normalized data.

Out of the three selected experiments, $\text{Ex}_1$ has the greater F1 score, which we consider to indicate a good balance between finding the maximum number of roundabout maneuvers (TPs) and minimizing false detections (FPs). Nevertheless, $\text{Ex}_1$ only detects 6 out of the 17 manually annotated passes and two of the roundabouts (RA1 and RA2) are never detected with it. On the other hand, $\text{Ex}_3$ is able to detect 14 of the passes at the expense of a higher number of false detections (54). In fact, $\text{Ex}_2$ is one of the 31\% of experiments to detect pass 2. $\text{Ex}_2$ is also able to detect most of the passes and only missing 5. However, unlike $\text{Ex}_3$, most of the predictions (63\%) using $\text{Ex}_2$ overlap with the manually annotated events.

Out of the 17 passes as depicted in Tab.~\ref{tab:results}, 5 correspond to roundabout maneuvers where the first exit was taken, and 5 to maneuvers where the second exit was taken. These passes are only detected by 15\% and 32\% of the experiment configurations, respectively. In contrast, maneuvers where the third exit was taken are detected by 60\% of the experiment configurations. This shows that the longer an actual passing maneuver through a roundabout takes, the higher is the maximum accuracy from all experiments as well as more detector configurations will spot such a maneuver. However, as depicted in Fig.~\ref{fig:results}, the exit taken does not necessarily affect the length of the maneuver, yet maneuvers that consist of a full turn (or more) are detected by more than 94\% of the experiment configurations, including $\text{Ex}_1$, $\text{Ex}_2$, and $\text{Ex}_3$. 

Another aspect to consider are multiple detections of a single roundabout. This can be seen in Fig.~\ref{fig:results} as multiple green boxes overlapping with the same pass in the manually annotated data. For instance, pass 15 was detected twice, and pass 17 was detected three times using $\text{Ex}_1$; and when using $\text{Ex}_2$ and $\text{Ex}_3$, each of the 4 passes that went through RA5 was detected twice. This raises the question as discussed in Sec.~\ref{sec:analysis-discussion} about whether a part of the roundabout in \gls{SFC} space is enough to claim that a roundabout maneuver was successfully detected. Moreover, experiment configurations that break down single maneuvers into multiple detections also have a higher number of false predictions as can be seen for $\text{Ex}_3$ in Fig.~\ref{fig:results}.

\begin{table*}[]
\centering
\resizebox{\textwidth}{!}{%
\begin{tabular}{cccccccccccccccccc}
\hline
Pass & 1 & 2 & 3 & 4 & 5 & 6 & 7 & 8 & 9 & 10 & 11 & 12 & 13 & 14 & 15 & 16 & 17 \\
\hline
Roundabout & RA1 & RA2 & RA3 & RA3 & RA4 & RA5 & RA4 & RA5 & RA6 & RA6 & RA7 & RA7 & RA7 & RA5 & RA4 & RA5 & RA6  \\
Exit taken & 3/4 & 2/4 & 6/5 & 2/5 & 2/4 & 4/4 & 1/4 & 2/4 & 1/4 & 1/4 & 1/4 & 4/4 & 1/4 & 2/4 & 8/4 & 5/4 & 14/4 \\
Hits (over all) & 60\% & 31\% & 96\% & - & 40\% & 81\% & 15\% & 46\% & 38\% & - & - & 85\% & 25\% & 42\% & 94\% & 96\% & 98\% \\
Max.~overlap~(\%) & 83\% & 83\% & 86\% & - & 82\% & 86\% & 69\% & 83\% & 82\% & - & - & 86\% & 83\% & 83\% & 86\% & 86\% & 86\% \\
\hline
\end{tabular}%
}
\caption{Boolean hits and maximum percentage of overlap across all experiments (omitting the geofence-based detector).}
\label{tab:results}
\end{table*}

The selected experiments described before therefore trigger three important discussions for Sec.~\ref{sec:analysis-discussion}: (i) the balance of metrics such as precision and recall to determine success, (ii) the trade-off between the data inputs and the amount of correctly identified passes, and (iii) whether fractions of a roundabout (small time-window), can be considered a TP as indicated above. 

\section{Analysis \& Discussion} \label{sec:analysis-discussion}


As expected, the automatic geofence-based detector, which exploits the GPS location to detect roundabouts, is scoring perfect results in terms of both precision and recall. However, as discussed before, GPS information may be considered as a \gls{GDPR}-sensitive information. An alternative detection strategy that does not rely on such information is therefore preferred.

\vspace{0.15cm}
\noindent\fbox{%
\begin{minipage}{.96\columnwidth}
\textbf{Answering RQ-1}, all experiment configurations were able to detect at least one roundabout passing maneuver, relying on non-\gls{GDPR}-sensitive data only. However, three passes were not detected by any of the experiment configurations.
\end{minipage}
}\\

According to the ordering as shown in Fig.~\ref{fig:results}, the most successful performing configuration is Morton using $\{a_x,a_y,\delta_S,\dot{\Psi}\}$ at \SI{5}{\hertz} (marked as $\text{Ex}_3$ in Fig.~\ref{fig:results}) returning 14 out of 17 maneuvers. Nevertheless, this high detection rate often comes at the expense of a low precision due to an increasing number of FPs. Moreover, the length in time of the detections is shorter than the manually annotated passes as the green ranges that are overlapping between the manual annotations and the experiment's results become smaller. Hence, in order to decide what experiment can be considered more successful over others is requiring a definition of a correctly identified passing maneuver (TP). The first option is Boolean detection, ie., caring about how many intersections with the ground truth are present. Therefore, the actual width of the green ranges is of secondary interest but rather the amount of distinct yet correct TPs is preferred. The second option is caring about which percentage of the manually annotated time windows corresponding to roundabout passes is detected by each of the experiments. 

\vspace{0.15cm}
\noindent\fbox{%
\begin{minipage}{.96\columnwidth}
\textbf{Answering RQ-2}, we discussed a number of experiment configurations that are able to detect roundabout passing maneuvers successfully. However, metrics for ``success'' might depend on the application of the proposed technique.
\end{minipage}
}\\

It is important to note that three passing maneuvers are not detected by any of the configurations listed Fig.~\ref{fig:results}: Passes 4, 10, and 11 as described in Tab.~\ref{tab:results}. These passes may be particularly challenging: In the former, the second exit was taken, which may be appearing an evasive maneuver; while in the latter the first exit was taken, making it similar to a simple right turn maneuver.

Experiment configurations that do not use steering wheel angle information like $\text{Ex}_1$ and $\text{Ex}_3$ detect on average a higher percentage of the manually annotated maneuvers. At the same time, configurations that do not use steering wheel angle information were only able to detect a maximum of 9 out of the 17 passes. Similarly, configurations that do not use the yaw rate to detect roundabout events also ranked higher yet none of them was able to detect passes 2, 4, 10, or 11. Interestingly, the passes which benefitted the most from not using steering wheel angle or yaw rate, were maneuvers when the roundabout was passed for at least one complete round.

In order to decide what experiment was more successful, the relative importance of TPs, FPs and FNs needs to be determined independently on whether we use Boolean or continuous definitions for correctly identified passes. In the context of \gls{ADAS/AD}, FPs are typically less problematic than FNs, so we ordered the experiments in Fig.~\ref{fig:results} based on the F1-score computed based on overlapping time windows. It is therefore important before using our technique to automatically detect maneuvers without relying on \gls{GDPR}-sensitive data to decide what is preferable: A small intersection with the ground truth might be enough to trigger a roundabout-specific \gls{ADAS/AD} behavior, ie., making the Boolean approach useful depending on such detections shall be post-processed. In a similar way, a different balance between TP, FP, and FN might be needed depending on the application context otherwise.


We report threats to validity following Feldt and Magazinius~\cite{FM10}. The data used in the experiments detailed in Section~\ref{fig:results} correspond to a limited data collection and showcase a selected number of events corresponding to roundabout pass-through maneuvers, which may pose an internal threat to validity. Even though the discussed input signals as reported in Tab.~\ref{tab:data-snowfox} are typically used for maneuver identification, we cannot claim generalizability of these results over alternative selections of signals or newly collected data at other locations; for example, many roundabouts along the trip included also a road bump to reduce speed, which on the one hand would even allow to improve the identification due to this particularity visible in $a_z$; however, this aspect is on the other not necessarily standard in other locations or countries and hence, we excluded this parameter. 


\section{Conclusion and Future Work}
\label{sec:conclusion-futurework}

We are presenting a systematic evaluation of \gls{SFC} to be used for automotive maneuver detection on the example of finding maneuvers for passing a roundabout. \glspl{SFC} allow for mapping multi-dimensional data into their single-dimensional representations and hence, such data can be processed computationally efficiently. This becomes important when working with ever-growing data lakes collected from large vehicle fleets, or when using automatic maneuver detectors that shall be used in real-time on the vehicle. 

Uniquely for our study is that we are interested in the performance of the maneuver detectors when dropping signals that may be \gls{GDPR}-sensitive such as GPS locations. Changing the sampling rate or the \gls{SFC} encoding can improve the number of true-positives but at an expense of finding more FPs as well, rendering such detectors impractical. In order to make decisions on which experiment configuration is more successful, application domain details need to be considered as discussed in Sec.~\ref{sec:results} using three examples with different time overlap and number of detections.

Future work may refine the design of the maneuver detectors further to lower the false-positive rates. In addition, the particularly challenging maneuver when leaving a roundabout already at the first exit in comparison to a plain turning right maneuver needs to be addressed further to better separate different automotive maneuvers when using non-\gls{GDPR}-sensitive signals. As we conducted our experiments with data from Gothenburg, Sweden, also the transferability of our findings needs to be explored further when looking at data from other regions.

\balance

\bibliographystyle{IEEEtran}
\bibliography{references.bib}

\end{document}